\documentclass[letterpaper, 10 pt, conference]{ieeeconf}  

\IEEEoverridecommandlockouts                              

\usepackage{amsmath}  
\usepackage{amssymb}  
\usepackage{verbatim} 
\usepackage{siunitx}  
\usepackage{hyperref} 
\usepackage[linesnumbered,ruled,vlined]{algorithm2e} 

\usepackage[nice]{nicefrac} 
\usepackage{subfig}
\usepackage{graphicx}

\usepackage[utf8]{inputenc}
\usepackage{pgfplots}
\DeclareUnicodeCharacter{2212}{−}
\usepgfplotslibrary{groupplots,dateplot}
\usetikzlibrary{patterns,shapes.arrows}
\pgfplotsset{compat=newest}

\title{\LARGE \bf
Data-Efficient Online Learning of Ball Placement in Robot Table Tennis
}

\author{Philip Tobuschat, Hao Ma, Dieter Büchler, Bernhard Schölkopf, Michael Muehlebach
\thanks{All authors are with the Max Planck Institute for Intelligent Systems, T{\"u}bingen (Germany)}
\thanks{}
\thanks{\hspace{-8pt}Preprint. Copyright 2023 by the author(s).}
}

\begin{document}

\maketitle
\thispagestyle{empty}
\pagestyle{empty}

\begin{abstract}

We present an implementation of an online optimization algorithm for hitting a predefined target when returning ping-pong balls with a table tennis robot. The online algorithm optimizes over so-called interception policies, which define the manner in which the robot arm intercepts the ball. In our case, these are composed of the state of the robot arm (position and velocity) at interception time. Gradient information is provided to the optimization algorithm via the mapping from the interception policy to the landing point of the ball on the table, which is approximated with a black-box and a grey-box approach. Our algorithm is applied to a robotic arm with four degrees of freedom that is driven by pneumatic artificial muscles. As a result, the robot arm is able to return the ball onto any predefined target on the table after about $2$-$5$ iterations. We highlight the robustness of our approach by showing rapid convergence with both the black-box and the grey-box gradients. In addition, the small number of iterations required to reach close proximity to the target also underlines the sample efficiency. A demonstration video can be found here: \url{https://youtu.be/VC3KJoCss0k}.

\end{abstract}


\section{Introduction}
\subsection{Background}
\label{sec:Background}
Reinforcement learning (RL) has proven to be a powerful and general framework that has many applications in robotics~\cite{Kober2013RoboticRLSurvey, Kormushev2010Archer, Zhu2018DexterousRL, Gullapalli1994RobotSkillsRL, Smart2002MobileRobotRL} and beyond~\cite{Li2017DeepRL}. However, modern RL approaches rely on deep neural networks~\cite{DeepRLAtari} and often require a large number of interactions with the environment until a reasonable policy is found. The search for a good policy is therefore often supported with numerical simulations which can be time-consuming and computationally expensive. 

Another downside is that the performance of RL algorithms is often sensitive to hyperparameter tuning, which requires extensive supervision by engineers, and limits their online applicability. Nevertheless, online adaptation is important for many real-world applications, where the environment or the robot dynamics may (gradually) change~\cite{Garcia2015SafeRLSurvey}.

In the following article, we propose an online optimization algorithm that mitigates some of these disadvantages. The algorithm uses gradient descent and allows for the incorporation of prior knowledge. We will evaluate our framework on a challenging robotic control task with a four-degrees-of-freedom robot arm driven by pneumatic artificial muscles (PAMs), see~\cite{DieterPamyIntroduction}. Our task is to enable the arm to return a ping-pong ball that is played to the robot back to any predefined target on the table. Our algorithm will iteratively update and learn the interception policy, where we take advantage of an already existing low-level controller that drives the robot arm, see~\cite{HaoILC}. We demonstrate that due to the inclusion of prior knowledge, our online optimization framework can be implemented in a data-efficient and robust manner. 

\subsection{Related Work} 
\label{sec:RelatedWork}
Most of the recent and competitive table tennis robots use an adaptation of RL for their control. The authors of~\cite{HuangDieterJointlyLearning} combine an RL algorithm that is trained to adjust the robot's joint trajectory with a regression model that predicts the ball's trajectory and the robot's joint trajectories. A combination of RL and imitation learning has been used by~\cite{GoogleGoalsEye} and~\cite{MuellingMotorPrimitives} and has shown strong results in terms of interception rates and landing point control, whereby~\cite{GoogleGoalsEye} explicitly evaluates performance for different target locations on the table. To facilitate and accelerate the learning process, \cite{DieterPAMYLearningFromScratch} developed a hybrid RL approach that simultaneously learns from a simulation and from executing a policy on the real-world system. The method is evaluated on a soft robot actuated by PAMs, which allows for safe exploration and thus removes the need for sophisticated policy initialization. All of these proposed methods use a high-dimensional action space that is searched through extensive trial and error, typically requiring thousands of iterations. 

Online optimization, which can be seen as an instance of online learning, was popularized by~\cite{ZinekevichOnline2003} and has mostly been used in theoretical settings to derive sample-complexity bounds, convergence guarantees, and convergence rates for learning algorithms. What makes the formulation of online optimization interesting is that it allows for unknown and even adversarial cost functions. This is in line with our setup, where the incoming ball trajectories are different from execution to execution and the system dynamics are assumed to be unknown and potentially slowly time-varying.

The soft robotics actuation principle of PAMs has seen application in a broad range of fields. Examples include exoskeletons for assistance and rehabilitation~\cite{Caldwell2007PAMExoskeleton, AlFahaam2018PAMExoskeleton, Beyl2014PAMExoskeleton, Park2014PAMExoskeleton, Zhang2008PAMExoskeleton}, legged locomotion~\cite{PAMCockroachInspiredRobot, Niiyama2010PAMAthlet, Vanderborght2006PAMBipedal, Lei2016PAMSpinningGait} and various aerospace applications~\cite{PAMAerospace}. A collection of different designs and additional application examples can be found in~\cite{PAMSurvey, PAMModeling}. Many first-principle models of PAMs have been built in order to facilitate their control, and a detailed summary of such modeling approaches is found in~\cite{PAMModeling}. Other control approaches that avoid utilizing sophisticated physical models include iterative learning control~\cite{HaoILC, Ai2020ILC} and RL~\cite{DieterPAMYLearningFromScratch}.

\subsection{Contribution} 
\label{sec:Contribution}
\begin{figure}[h]
\centering
\includegraphics[width = 0.8\linewidth]{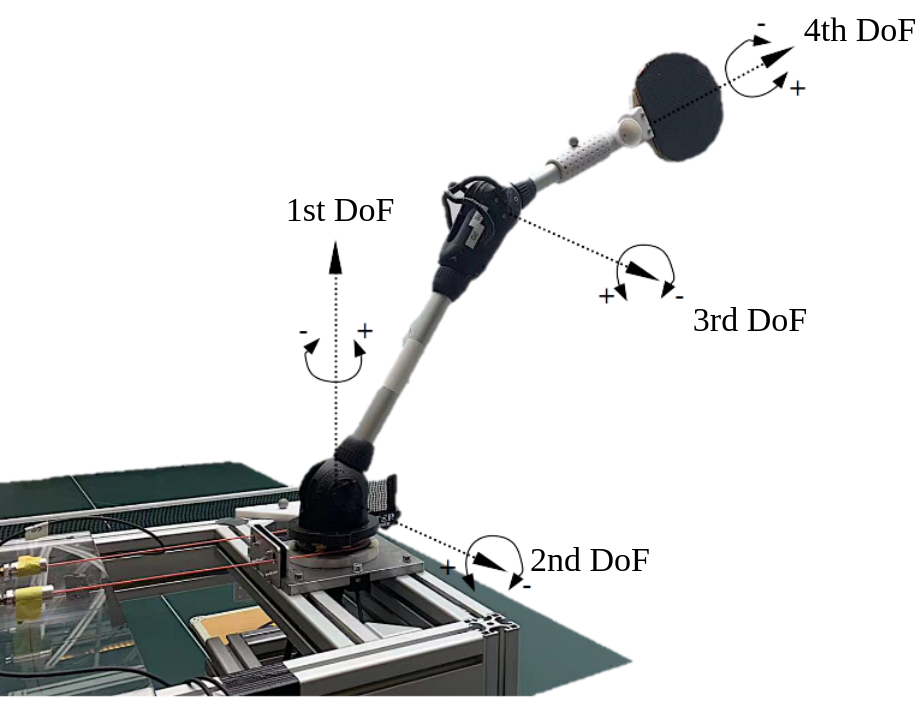}
\caption{The figure shows the structure of the robot arm with its four rotational degrees of freedom (DoFs). Each joint is actuated by a pair of PAMs.}
\label{fig:main_body}
\end{figure}

The main contribution of our work is to propose a robust and data-efficient online optimization algorithm for solving learning tasks that arise in robotic systems. This contrasts traditional applications of online optimization algorithms, which mainly consist of machine learning applications such as recommendation systems, packet routing, or spam filtering~\cite{HazanOCO}. More precisely, we apply our framework to a four-degrees-of-freedom robot arm that is driven by PAMs and is illustrated in Figure~\ref{fig:main_body}. The algorithm takes advantage of an existing low-level controller and is able to return balls to predefined targets with an accuracy of about \SI{25}{\cm}. An important aspect of our framework is to build a mapping that connects the interception policy, which is defined as the state of the robot arm at interception time, to the Euclidean distance between the target and the landing point of the ball. The distance between the target and the landing point defines the loss in the online optimization and is measured by a vision system~\cite{Vision_Gomez_Gonzalez_2019}. At each iteration, i.e., after each returned ball, the interception policy will be updated online by a gradient descent method, where the gradients are approximated with either a black-box or a grey-box approach.

We note that the lack of gradient information due to the unknown underlying system dynamics is a key bottleneck, not only when applying online optimization algorithms to robotic systems, but also for RL in general. The traditional approach of using (stochastic) finite differences is often biased and has a high variance, which considerably slow down the optimization. Therefore, an important contribution of our work is to propose both a grey-box and a deep learning approach that approximate the mapping from the interception policy to the loss. We compare the results when employing either of the two mappings and find that in both cases the distance error converges toward zero (on average) at similar rates and with a similar variance.

Compared to existing RL algorithms for robotic systems such as Q-learning or actor-critic methods, our framework is very data efficient. In our extensive experiments that include different initial policies and different targets, our algorithm required only about $2$-$5$ iterations to hit a target with an error below \SI{25}{\cm}. The algorithm is therefore directly applied to the real-world robotic system and does not require a simulator (although a simulator could be used to warm-start the online optimization, which would further reduce the number of iterations needed).

\subsection{Structure}
\label{sec:Structure}
This article is structured as follows: In Section~\ref{sec:OnlineOptimization}, we formulate the task of finding interception policies as an online optimization problem and describe the gradient descent algorithm that we use subsequently. In Section~\ref{sec:PredictionModels}, we describe two fundamentally different approaches for predicting the landing points (either grey-box or black-box), which will provide important gradient information to the optimization algorithm. The performance of both approaches will be evaluated in real-world experiments as described in Section~\ref{sec:Result}. The article concludes with a summary in Section~\ref{sec:Conclusion}.
\section{ONLINE OPTIMIZATION}
\label{sec:OnlineOptimization}

\subsection{Problem Formulation}
\label{sec:ProblemFormulation}
In our work, the robot arm learns to hit and return a ball shot from a launcher (presented in~\cite{Dittrich2022AIMY}) onto a predefined target $r_{\text{target}} \in \mathbb{R}^2$ on the table. We ensure that the $z$-coordinate of the global coordinate system stands orthogonal to the table, and unless otherwise specified, all the following calculations take place in the global coordinate system. The target is then within the $x$-$y$ plane at table height, that is, $r_{\text{target}}=\left(x_{\text{target}}, y_{\text{target}}\right)^{\text{T}}$. Our approach can easily be extended to aim at a three-dimensional target in space. 

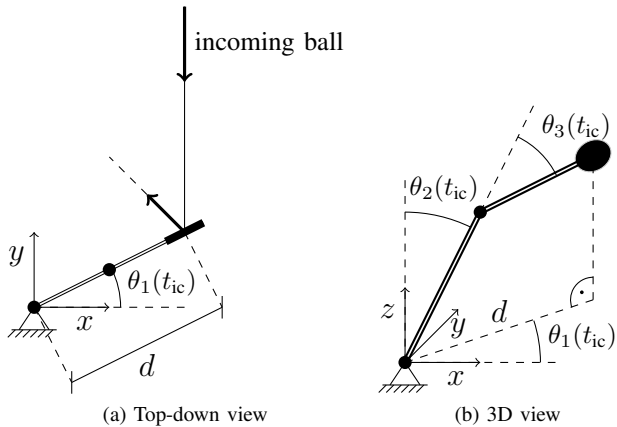
\begin{figure}[h]
    \centering
    \subfloat[Top-down view]{\centering
    \begin{tikzpicture}

    \draw (0.0, 4.0) -- (0.0, 1.0);
    \draw[->, line width=1.2] (0.0, 4.0) -- (0.0, 3.0) node[midway, right] {incoming ball};
    \draw[->, line width=1.2] (0.0, 1.0) -- (-0.5, 1.5);
    \draw[dashed] (-0.5, 1.5) -- (-1.0, 2.0);

    \draw[double] (-2.0, 0.0) -- (0.25, 1.125);
    \draw[line width = 3] (-0.25, 0.875) -- (0.25, 1.125);
    \fill (-2.0, 0.0) circle (2.5pt);
    \fill (-1.0, 0.5) circle (2.5pt);

    \draw (-0.85, 0.0) arc (0:26.57:1.15) node[midway,right, yshift=2pt] {$\theta_1(t_{\text{ic}})$};
    \draw[dashed] (-2.0, 0.0) -- (-0.0, 0.0);

    \draw[->] (-2.0, 0.0) -- (-1.0, 0.0) node[midway, below, xshift=5pt] {\large $x$};
    \draw[->] (-2.0, 0.0) -- (-2.0, 1.0) node[midway, left, yshift=5pt] {\large $y$};

    \draw (-2.0-0.2, 0.0-0.3) -- (-2.0, 0.0);
    \draw (-2.0+0.2, 0.0-0.3) -- (-2.0, 0.0);
    \draw (-2.0-0.3, -0.3) -- (-2.0+0.3, -0.3);
    \foreach \x in {0,...,5}
        \draw (-2.0-0.25 +\x * 0.1, -0.3) -- (-2.0-0.25-0.1 + \x*0.1, -0.4);

    \draw[dashed] (-2.0, 0.0) -- (-1.5, -1.0);
    \draw[dashed] (0.0, 1.0) -- (0.5, 0.0);
    \draw (-1.5, -1.0) -- (0.5, 0.0) node[midway, below] {\large $d$} ;
    \draw (-1.5, -1.0-0.15) -- (-1.5, -1.0+0.15);
    \draw (0.5, 0.0-0.15) -- (0.5, 0.0+0.15);

\end{tikzpicture}
    \label{fig:TopDowdSketch}}
    \subfloat[3D view]{\centering
    \begin{tikzpicture}

    \draw (-0.0-0.2, 0.0-0.3) -- (-0.0, 0.0);
    \draw (-0.0+0.2, 0.0-0.3) -- (-0.0, 0.0);
    \draw (-0.0-0.3, -0.3) -- (-0.0+0.3, -0.3);
    \foreach \x in {0,...,5}
        \draw (-0.0-0.25 +\x * 0.1, -0.3) -- (-0.0-0.25-0.1 + \x*0.1, -0.4);

    \draw[double, thick] (0, 0) -- (1, 2);
    \draw[double, thick] (1, 2) -- (2.5, 2.75);
    \fill (0.0, 0.0) circle (2.5pt);
    \fill (1, 2) circle (2.5pt);
    \fill (2.5, 2.75) circle (2.5pt);
    \draw[draw=black,draw opacity=0.2, fill=black, fill opacity=1, rotate around={30:(2.5, 2.75)}] (2.5, 2.75) ellipse (0.25 and 0.2);

    \draw[dashed] (0.0, 0.0) -- (2.0, 0.0);
    \draw (1.8, 0.0) arc (0:18:1.8) node[midway,right, yshift=2pt] {$\theta_1 (t_{\text{ic}})$};

    \draw[dashed] (0, 0) -- (0, 2.5);
    \draw (0.0, 2) arc (90:64:2) node[midway,above, yshift=2pt, xshift=2pt] {$\theta_2 (t_{\text{ic}})$};
    
    \draw[dashed] (1, 2) -- (1.75, 3.5);
    \draw (1+1.5/3, 2+3/3) arc (64:26:1.5/3*5^0.5) node[midway,right, xshift=-3pt, yshift=10pt] {$\theta_3 (t_{\text{ic}})$};

    \draw[->] (0.0, 0.0) -- (1.0, 0.0) node[midway, below, xshift=5pt] {\large $x$};
    \draw[->] (0.0, 0.0) -- (0.5*2^0.5, 0.5*2^0.5) node[below, yshift=0pt, xshift=0pt] {\large $y$};
    \draw[->] (0.0, 0.0) -- (0.0, 1.0) node[midway, left, yshift=5pt] {\large $z$};

    \draw[dashed] (0.0, 0.0) -- (2.5, 5/6) node[midway, above] {\large $d$};
    \draw[dashed] (2.5, 3-1/6) -- (2.5, 5/6);

    \draw (2.5, 1.3-1/6) arc (90:200:0.3);
    \fill (3-0.15-0.5, 1.1-1/6) circle (0.7pt);

\end{tikzpicture}
    \label{fig:3DSketch}}
    \caption{The figure shows a sketch of the robot arm. Subigure (a) illustrates how $\theta_1$ defines the interception point and subfigure (b) shows how this subsequently fixes $\theta_2$ and $\theta_3$ as well.}
    \label{fig:Sketch}
\end{figure}

In the following, $\theta_j$ and $\dot{\theta}_j$ denote the angle and angular velocity of degree of freedom $j \in \{ 1, 2, 3, 4\}$. The landing point $ r_{\text{landing}} \in \mathbb{R}^2$ after the ball is returned, also within the $x$-$y$ plane at the table height, is dependent on the interception policy $\phi \in \mathcal{K} \subseteq \mathbb{R}^{2}$, where $\mathcal{K}$ denotes the feasible set as described below. As illustrated in Figure~\ref{fig:Sketch}, for a given ball trajectory the angle $\theta_1$ uniquely determines the interception time point, $t_\text{ic}$, as well as the angles $\theta_2$ and $\theta_3$ of the remaining degrees of freedom at time $t_\text{ic}$. At the interception time point, we fix the velocity $\dot{\theta}_1 =$ \SI{6}{\radian\per\second} and set the velocities $\dot{\theta}_2$, $\dot{\theta}_3$, and $\dot{\theta}_4$ to zero (this amounts to a straight return that does not induce extra spin). The algorithm works in the same way when $\dot{\theta}_1$, $\dot{\theta}_2$, or $\dot{\theta}_3$ are set to different values. Our interception policy therefore controls the interception time point and the angle of the return with $\theta_1$ and the distance of the return with $\theta_4$. We also conducted experiments where both $\theta_4$ and $\dot{\theta}_1$ were used to control the distance, but found that $\theta_4$ was the dominant factor, which led us to the choice
\begin{equation*}
    \phi = \left(\theta_1(t_\text{ic}), \theta_4(t_\text{ic}) \right)^{\text{T}}.
\end{equation*}
The feasible set $\mathcal{K}$ is introduced to ensure that the interception policy does not exceed the physical limits of the robot arm and is chosen as $\mathcal{K} = \{ \phi \in \mathbb{R}^{2}| -\frac{\pi}{2} \leq \theta_1(t_\text{ic}) \leq \frac{\pi}{2}, -\frac{\pi}{4} \leq \theta_4(t_\text{ic}) \leq \frac{\pi}{4}\}$. The set is convex and thanks to the high compliance and power-to-weight ratio of the PAMs~\cite{DieterPamyIntroduction}, it is large enough to cover the entire range of interception policies needed in practice.
In the following, we assume that $r_{\text{landing}}$ can be divided into a deterministic part, $\tilde{r}_{\text{landing}}$, and an additive disturbance as follows:
\begin{equation*}
    r_{\text{landing}} \left( \phi, n \right) = \tilde{r}_{\text{landing}} \left( \phi \right) + n,
\end{equation*}
where the disturbance is non-repetitive and captures the fluctuation of the different incoming ball trajectories and the process noise of the robot arm. We note that according to the online learning/online optimization framework,~\cite{HazanOCO}, the disturbance is not necessarily required to be stochastic and zero mean for our framework to perform well.

For any fixed target, the learning task will be formulated as the following online optimization problem:
\begin{align}
\begin{split}
   \label{eq:loss}
   &\sum_{i=1}^{N} L^{i} (\phi^{i} ) \rightarrow \text{min},\\
   &\text{where } \phi^{i} \in \mathcal{K}, L^{i} (\phi^{i}) = \frac{1}{2} | r_\text{landing}(\phi^{i}, n^{i}) - r_\text{target} |^2, \\
   &\text{for all } i=1,\dots,N,
\end{split}
\end{align}
where $N$ denotes the number of iterations (not necessarily fixed). We denote by $|\cdot|$ the Euclidean distance and by the superscript $\left(\cdot\right)^{i}$ the iteration number. The non-repetitive disturbance $n^{i}$ leads to an iteration-dependent loss function with the interception policy as its only argument. We note that at the iteration $i$ the optimization algorithm has only access to the data from previous iterations $l = i-1, \dots, 1$ while the disturbance $n^{i}$ is unknown.

\subsection{Online Gradient Descent}
\label{sec:OnlineGD}
We optimize~\eqref{eq:loss} by applying online gradient descent, where according to the chain rule, the gradient of the loss function at each iteration is calculated as follows:
\begin{equation*}
\nabla L^{i}(\phi^{i})= \underbrace{\left( \frac{\partial \tilde{r}_\mathrm{landing} \left(\phi^{i}\right) }{\partial \phi} \right)^{\mathrm{T}}}_{\text{Part 1}}  \underbrace{(r_\mathrm{landing}\left( \phi^{i}, n^{i} \right) - r_\mathrm{target})}_{\text{Part 2}}, 
\end{equation*}
with $i=1,\dots,N$. We note that the value of Part $2$ can be easily obtained because $r_{\text{target}}$ is predefined and fixed, and $r_{\text{landing}}\left( \phi^{i}, n^{i}\right)$ at the $i$-th iteration is observed by the vision system. In contrast, the value of Part $1$ is more difficult to determine, since the mapping from interception policies to landing points is unknown a priori. To resolve this issue, we will use either a grey-box or a black-box approach for predicting the landing point as a function of $\phi$, which will be denoted by $g(\phi)$. This provides us with the following approximations:
\begin{equation*}
    \tilde{r}_{\text{landing}} \left( \phi \right) \approx g\left(\phi\right),
    \quad \frac{\partial \tilde{r}_{\text{landing}} \left( \phi \right)}{\partial \phi} \approx \frac{\partial g(\phi)}{ \partial \phi}.
\end{equation*}
The grey-box and the black-box models will be discussed in detail in Section~\ref{sec:PredictionModels}.

The implementation of our online gradient descent scheme is summarized in Algorithm~\ref{algo:OnlineOpt}. We note that $\Pi_{\mathcal{K}}$ denotes the projection onto the feasible set, which can be evaluated in closed form. {\sc Interception} denotes the physical process of ejecting a ball from the ball launcher, intercepting it with the racket in a manner defined by the current policy, and observing the landing point on the table. 

An important difference compared to the online optimization formulation in~\cite{HazanOCO} is that we only rely on an approximation of the gradient, which contains a part that can be measured directly (Part 2) and a part that is modeled either from first principles or in a data-driven manner. This gradient approximation is a key innovation in our work and is the main driving force that makes our approach data efficient.
\SetKwInOut{Given}{Given}
\begin{algorithm}
\SetAlgoLined
\setcounter{AlgoLine}{0}
\Given{target $r_{\mathrm{target}}$, $N$, step lengths $ \left\{ \alpha^i \right\}^{N}_{i=1}$}
\KwIn{initial policy $\phi^1$}
  \For{$i \gets 1$ \KwTo $N$}{
    $r_{\mathrm{landing}}^{i} \gets$ {\sc Interception}($\phi^i$) \;
    $\widetilde{\phi}^{i+1} \gets \phi^i - \alpha^i \nicefrac{\partial g(\phi^{i})}{\partial \phi}^{\mathrm{T}} \left(r_{\mathrm{landing}}^{i} - r_{\mathrm{target}}\right)$ \;
    $\phi^{i+1} \gets \Pi_{\mathcal{K}} \left( \widetilde{\phi}^{i+1} \right)$;
}
\caption{{\sc Approximate Online Gradient Descent}}
\label{algo:OnlineOpt}
\end{algorithm}

\section{Landing Point Prediction Model}
\label{sec:PredictionModels}
As mentioned in the previous section, an approximate mapping from the interception policy $\phi$ to the landing point is required to provide gradient information to our online learning framework. In this section, we discuss a grey-box and a black-box approach for approximating this mapping. 

\subsection{Grey-Box Model}
We use first principles to predict the landing point as a function of $\phi$, which includes an impact model between the ball and the racket as well as a model of the ball in free flight. First of all, we introduce the position vector $p[k] \in \mathbb{R}^3$ and the velocity vector $v[k] \in \mathbb{R}^3$ of the ball as discrete variables parameterized by the index $k$ that captures the time evolution,
\begin{align*}
    \begin{split}
        p[k] &= \left(p_x[k], p_y[k], p_z[k]\right)^{\text{T}},\\
        v[k] &= \left(v_x[k], v_y[k], v_z[k]\right)^{\text{T}}.
    \end{split}
\end{align*}

The impact with the racket is assumed to be an instantaneous event, where the variables right before and after the impact are denoted by $(\cdot)^{-}$  and $(\cdot)^{+}$, respectively. The rotation matrix that describes the orientation of the racket with respect to its rest configuration is denoted by $\Gamma(\phi) \in \mathbb{R}^{3 \times 3}$. The racket velocity expressed in the global coordinate frame is denoted by $v_\mathrm{R}(\phi) \in \mathbb{R}^{3}$. Both of these variables are derived from the kinematics of the robot arm. Thus, the racket impact model can be summarized as
\begin{align}
\begin{split}
    p^{+} &= p^{-}, \\
    v^{+}(\phi) &= \Gamma\left(\phi\right) M_{\mathrm{R}} \Gamma\left(\phi\right)^{\text{T}} \underbrace{\left(v^{-} - v_{\mathrm{R}} \left(\phi\right) \right)}_{\text{relative velocity}} + v_{\mathrm{R}} \left(\phi\right),
    \label{eq:RacketImpact}
\end{split}
\end{align}
where $M_\mathrm{R} = \text{diag} \left(0.75, -0.75, 0.75\right)$ is a linear impact model. The negative component in $M_\mathrm{R}$ is aligned with the normal direction of the racket's surface in its rest position.

For modeling the free flight of the ball we introduce $\xi[k] \in \mathbb{R}^{6}$, which combines the position and velocity, i.e., $\xi[k]^{\text{T}} = \left(p[k]^{\text{T}}, v[k]^{\text{T}}\right)$. In principle, the full state of the ball also includes the ball's spin. However, the spin influences the free-flight motion of the ball only slightly through the Magnus effect~\cite{Huang2011TrajectoryPrediction}. The purpose of our model is to compute gradients for our online optimization algorithm, and we therefore favor low complexity over detailed modeling and omit the ball's spin. As a result, the free-flight dynamics are approximated as follows: 
\begin{multline}
    \xi[k+1] = q\left(\xi[k]\right)\\
             = \xi[k] + T_\text{s} \left(v[k]^{\text{T}}, - k_\mathrm{D} |v[k]| v[k]^{\text{T}} +g_\text{0}^{\text{T}}\right)^{\text{T}},
    \label{eq:DiscreteDynamics}
\end{multline}
where $k_\mathrm{D}=\SI{0.106}{\per \meter}$ denotes the drag coefficient and $g_\text{0}=(0, 0, -\SI{9.8}{\meter \per\square\second})^{\text{T}}$ denotes the gravitational acceleration. We write $T_\text{s}$ for the time step. The initial condition for~\eqref{eq:DiscreteDynamics} is given by the ball's state after the impact with the racket, that is, $p[0]=p^{+}$, $v[0]=v^{+}(\phi)$, and $\xi[0] = \xi^{+}(\phi)$.

The evolution stops once the predicted trajectory intersects the $x$-$y$ plane at the table's height, which is denoted as $p_{z, \mathrm{table}}$. At each index $k$, we predict the remaining time until the ball reaches the table height based on a simplified evolution of the ball's $z$-coordinate, which neglects aerodynamic drag, as
\begin{equation}
    T_{\mathrm{r}}\left(\xi\left[k\right]\right) = \frac{v_z[k]}{\hat{g}} + \sqrt{\left( \frac{v_z[k]}{\hat{g}} \right)^2 + 2\cdot \frac{p_z[k] - p_{z, \mathrm{table}}}{\hat{g}}},
    \label{eq:T_remain}
\end{equation}
where $\hat{g} = \SI{9.8}{\meter \per\square\second}$. Once $T_\mathrm{r}\left(\xi\left[k\right]\right)$ is less than or equal to $T_\mathrm{s}$, we perform the next and last step with the shortened step length of $T_\mathrm{r}\left(\xi\left[k\right]\right)$ instead of $T_\mathrm{s}$ and terminate the evolution. The last index $k$ before this shortened step will be referred to as $k_{\mathrm{max}}$, and we refer to the last time step as $T_\mathrm{last}=T_\mathrm{r}\left(\xi\left[k_{\text{max}}\right]\right)$. 

Thus, by combining the racket impact model and the free-flight model, we can predict the landing point $\tilde{r}_\mathrm{landing}$. This concludes the landing point prediction model.

Next, we will show how to calculate the required gradient $\nicefrac{\partial \tilde{r}_\mathrm{landing}}{\partial \phi}$, which is naturally included in $\nicefrac{\partial \xi_\mathrm{landing}}{\partial \phi}$. We apply the chain rule and conclude
\begin{equation*}
    \frac{\partial \xi_\mathrm{landing}}{\partial \phi} = \frac{\partial \xi_\mathrm{landing}}{\partial \xi[k_\mathrm{max}]} \frac{\partial \xi [k_\mathrm{max}]}{\partial \xi^{+}} \frac{\partial \xi^{+}}{\partial \phi}.
\end{equation*}

We will evaluate each of the three terms individually, starting with $\nicefrac{\partial \xi^{+}}{\partial \phi}$. Deriving~\eqref{eq:RacketImpact} with respect to $\phi$ yields
\begin{multline*}
    \frac{\partial v^{+}}{\partial \phi} = \left( \frac{\partial \Gamma}{\partial \phi} M_{\mathrm{R}} \Gamma^{\text{T}} - \Gamma M_{\mathrm{R}} \frac{\partial \Gamma}{\partial \phi}^{\text{T}} \right) (v^{-} - v_{\text{R}}) \\
     - \Gamma M_{\mathrm{R}} \Gamma^{\text{T}} \frac{\partial v_{\mathrm{R}}}{\partial \phi} + \frac{\partial v_{\mathrm{R}}}{\partial \phi},
\end{multline*}
where the terms $\nicefrac{\partial \Gamma}{\partial \phi}$ and $\nicefrac{\partial V_{\mathrm{R}}}{\partial \phi}$ can be computed based on the system's kinematics.

For the term $\nicefrac{\partial \xi[k_\mathrm{max}]}{\partial \xi^{+}}$, we recall that $\xi[k_\mathrm{max}]$ is calculated by repeatedly applying the evolution function $q\left(\xi[k]\right)$ with a constant time step starting from $\xi^{+}$. For each index $k$ we apply the function $q$, which yields
\begin{equation*}
\frac{\partial \xi[k]}{\partial \xi[k-1]} =\left. \frac{\partial q(\xi)}{\partial \xi} \right|_{\xi=\xi[k-1]}, k = 1, \dots, k_\mathrm{max},
\end{equation*}
where $\nicefrac{\partial q(\xi)}{\partial \xi}$ can be directly calculated based on~\eqref{eq:DiscreteDynamics}. We get
\begin{equation*}
    \frac{\partial \xi[k_\mathrm{max}]}{\partial \xi^{+}} = \prod_{k = 1}^{k_\mathrm{max}} \left. \frac{\partial q(\xi)}{\partial \xi} \right|_{\xi = \xi[k-1]}. 
\end{equation*}
For evaluating the last term $\nicefrac{\text{d} \xi_{\mathrm{landing}}}{\text{d} \xi[k_{\text{max}}]}$, we recall the functional dependence of the last evolution before the table impact as $\xi_{\mathrm{landing}} = \xi_{\mathrm{landing}}\left(\xi[k_\mathrm{max}], T_{\mathrm{last}}\right)$ and expand $\nicefrac{\partial \xi_{\mathrm{landing}}}{\partial \xi[k_\mathrm{max}]}$ as follows:
\begin{equation*}
    \frac{\partial \xi_{\mathrm{landing}}}{\partial \xi[k_\mathrm{max}]} = \frac{\partial \xi_{\mathrm{landing}}}{\partial \xi[k_\mathrm{max}]} + \frac{\partial \xi_{\mathrm{landing}}}{\partial T_{\mathrm{last}}} \frac{\partial T_{\mathrm{last}}}{\partial \xi[k_\mathrm{max}]},
\end{equation*}
where $\nicefrac{\partial \xi_{\mathrm{landing}}}{\partial \xi[k_\mathrm{max}]}$ and $\nicefrac{\partial \xi_{\mathrm{landing}}}{\partial T_{\mathrm{last}}}$ can be rewritten as $\nicefrac{\partial q(\xi)}{\partial \xi}$ and $\nicefrac{\partial q(\xi)}{\partial T_{\mathrm{s}}}$ evaluated at $\xi = \xi[k_\mathrm{max}]$ and $T_\mathrm{s} = T_{\mathrm{last}}$, respectively. The remaining term $\nicefrac{\partial T_{\mathrm{last}}}{\partial \xi[k_{\text{max}}]}$ is directly calculated from \eqref{eq:T_remain}.

\subsection{Black-Box Model}
We use a black-box model as an alternative approach for predicting the landing point. More precisely, we employ a neural network $g_{\text{NN}}$ with only fully connected layers, which predicts the landing point on the table based on a given interception policy, that is, $\tilde{r}_\mathrm{landing} = g_\mathrm{NN}(\phi)$. The architecture consists of four hidden layers, each of them with four nodes, and the hyperbolic tangent function is used as the activation function for all layers except for the output layer.
We collected a dataset of about $3000$ data points that cover the relevant range of $\phi \in \mathcal{K}$, where each data point corresponds to a successful return with the robot arm (if the ball only hit the edge of the racket or the robot missed the ball, the trajectory was discarded).
We trained the neural network using \verb|pytorch|, which required about $500$ epochs with the \verb|Adam| optimizer~\cite{Adam}. The gradient of the landing point with respect to the interception policy is then computed with \verb|pytorch|.
\section{Results}
\label{sec:Result}

We will use the following metrics for analyzing the performance of our online optimization algorithm:
\begin{align}
\begin{split}
\bar{r}_i &= \frac{1}{i} \sum_{j=1}^i r_{\mathrm{landing}}^{j},\quad
\epsilon_i = \left |  r_{\mathrm{target}} - \bar{r}_i \right |,
\\
\sigma_i &= \sqrt{\frac{1}{i} \sum_{j=1}^i \left | r_{\mathrm{landing}}^{j} - \bar{r}_i \right |^2 },\quad i=1,\dots,N,
\label{eq:PerformanceMetrics}
\end{split}
\end{align}
where $\bar{r}_i$ denotes the mean landing point for the first $i$ iterations and $\epsilon_i$ denotes the distance error between the target and the mean landing point. The standard deviation of landing points up to iteration $i$ is denoted by $\sigma_i$.

\begin{figure}[h]
    \centering
    \vspace*{6pt}
    \input{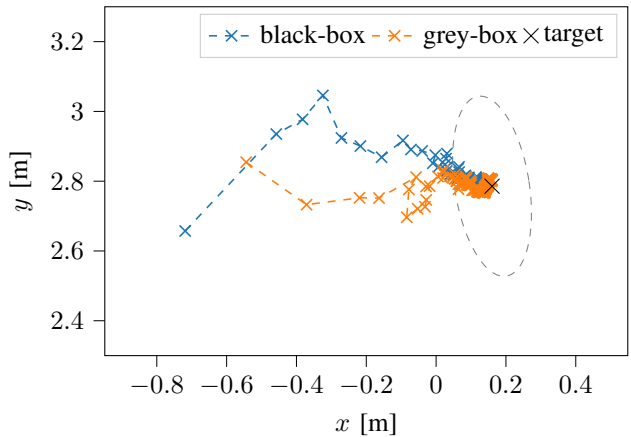}
    \caption{The figure shows the evolution of the mean landing points $\bar{r}_i$ on the $x$-$y$ plane at the height of the table. The evolution using the grey-box method is denoted by orange crosses, while the evolution using the black-box method is denoted by blue crosses. The grey dashed line provides an estimate of the standard deviation. We note that the convergence appears to be slower than it actually is due to the fact that we plot $\bar{r}_i$.}
    \label{fig:MeanLandingPoints}
\end{figure}

Furthermore, we experimentally determine the inherent variance of landing points (due to the process noise and the fluctuations in the incoming ball trajectories) by running three experiments over $200$ iterations each with different reasonable interception policies that remain fixed. We find that the inherent standard deviation of landing points for a fixed policy is around $\underline{\sigma} = \SI{25}{\cm}$. The standard deviation differs quite significantly along the two axes as shown in Figure~\ref{fig:MeanLandingPoints} (grey ellipsoid).

In the following, we will show the results of different experiments that characterize the convergence and robustness properties of the algorithm. For all of the following experiments, step lengths $\alpha^i = \nicefrac{\alpha^1}{\sqrt{i}},i=1,\dots,N$ are employed in Algorithm~\ref{algo:OnlineOpt}.

\subsection{Robustness}
\label{sec:ResultsConvergence}
In this section, we test the robustness of the online optimization framework by comparing results produced by either employing the grey-box or the black-box landing point prediction for approximating $\nicefrac{\partial g}{\partial \phi}$.

In Figure~\ref{fig:MeanLandingPoints}, we show the results from two $200$-iteration experiments with the grey-box or the black-box approach. The figure shows the evolution of the mean landing points $\bar{r}_i$ within the $x$-$y$ plane. We notice that although the evolution directions of the two methods differ, they both converge to the predefined target point.
\begin{figure}[h]
    \centering
    \vspace*{6pt}
    \subfloat[Mean distance error]{\centering
    \input{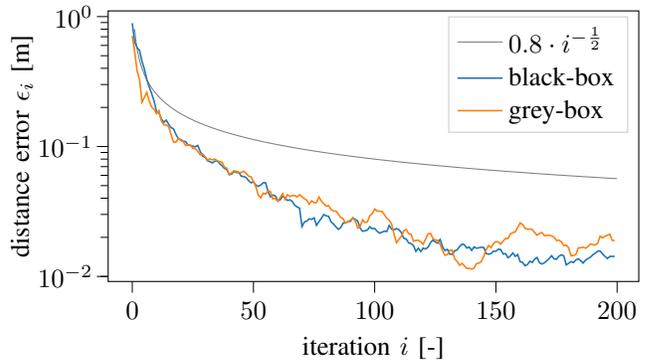}
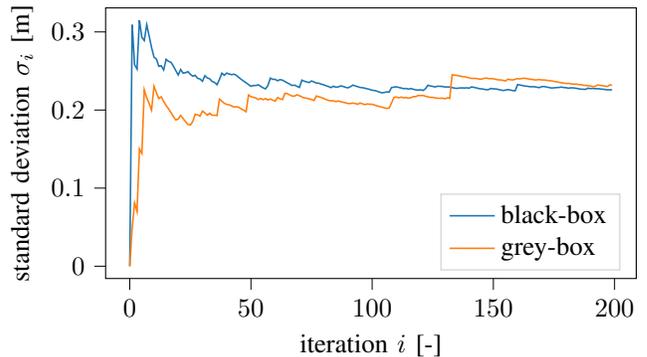
    \label{fig:MeanDistanceError}}\\
    \subfloat[Standard deviation]{\centering
\begin{tikzpicture}

\definecolor{darkgray176}{RGB}{176,176,176}
\definecolor{darkorange25512714}{RGB}{255,127,14}
\definecolor{lightgray204}{RGB}{204,204,204}
\definecolor{steelblue31119180}{RGB}{31,119,180}

\begin{axis}[
height=0.6\linewidth,
width =\linewidth,
legend pos = south east,
legend cell align={left},
legend style={fill opacity=0.8, draw opacity=1, text opacity=1, draw=lightgray204},
tick align=outside,
tick pos=left,
x grid style={darkgray176},
xlabel={iteration \(\displaystyle i\) [-]},
xmin=-9.95, xmax=208.95,
xtick style={color=black},
y grid style={darkgray176},
ylabel={standard deviation \(\displaystyle \sigma_i\) [m]},
ymin=-0.0157496809721493, ymax=0.330743300415135,
ytick style={color=black}
]
\addplot [semithick, steelblue31119180]
table {%
0 0
1 0.309124548010982
2 0.258060169130077
3 0.252530290395285
4 0.314993619442986
5 0.293035625293659
6 0.288982575303917
7 0.309097869491614
8 0.294837992209416
9 0.27993773880761
10 0.267779410513225
11 0.265241248078486
12 0.255966669982255
13 0.256655361462731
14 0.251330086432839
15 0.264891680412309
16 0.262368487634854
17 0.261087851898129
18 0.256238242410603
19 0.250407827165389
20 0.244852131144475
21 0.251909040887385
22 0.247228783833666
23 0.247759644679393
24 0.249172719588177
25 0.245657290213973
26 0.243332820084683
27 0.244421484145323
28 0.240555950970608
29 0.239749284260074
30 0.236799015905897
31 0.244010692726003
32 0.240980465784106
33 0.239970282172531
34 0.236518135285807
35 0.235210677354716
36 0.232365924540934
37 0.23916844735617
38 0.247167655311101
39 0.244535265332
40 0.247306728344025
41 0.246142094233033
42 0.245280300763381
43 0.245862354192301
44 0.244341916333953
45 0.241674822320142
46 0.239246206897095
47 0.236980475300456
48 0.234916849555558
49 0.23255839897254
50 0.23036673607133
51 0.231445498984365
52 0.231335657688642
53 0.232294364622541
54 0.23038345943901
55 0.228385670891608
56 0.227074156724471
57 0.231439965231487
58 0.240851335520352
59 0.239334103974261
60 0.239230978080606
61 0.237449361990834
62 0.238961470232347
63 0.238190110102444
64 0.236499364646613
65 0.234777177477642
66 0.233138393338556
67 0.231434485090054
68 0.231830090475329
69 0.230308390290958
70 0.228880592241676
71 0.238259180939091
72 0.236862448032139
73 0.236027266493218
74 0.235124429242348
75 0.237183994812889
76 0.235936245867026
77 0.234518187764656
78 0.234314909399957
79 0.233064624412911
80 0.232384228433415
81 0.231052733919227
82 0.230058554358971
83 0.228685750994577
84 0.230513645406199
85 0.229217279724817
86 0.2295834410349
87 0.230829230849945
88 0.230273252111623
89 0.22907048866704
90 0.228377958523437
91 0.229777016404644
92 0.231892995030276
93 0.230877532775136
94 0.229941486766881
95 0.228889904511641
96 0.228607311261886
97 0.227905282910242
98 0.227174964341007
99 0.226344171081575
100 0.225330439168831
101 0.224991488652584
102 0.224031506487045
103 0.223002373861946
104 0.222035989967809
105 0.222415414054524
106 0.223221646699135
107 0.222854339178071
108 0.228772537966495
109 0.229283160826056
110 0.229717200888776
111 0.228837540547469
112 0.22817226720129
113 0.227354664668955
114 0.227611990948548
115 0.226697908162705
116 0.227529633384208
117 0.226579840234897
118 0.225771566356433
119 0.225536466677772
120 0.225891971611674
121 0.22499435500919
122 0.227269247536485
123 0.231692631328068
124 0.230791636962078
125 0.230247819575293
126 0.229515628504013
127 0.230515945533676
128 0.230348714735489
129 0.230137416950622
130 0.229460294979751
131 0.229545630029951
132 0.229366462212083
133 0.228734627873104
134 0.227939344223252
135 0.227755518277157
136 0.22814585225427
137 0.228366758327526
138 0.228126512560624
139 0.228425771333972
140 0.227952869700803
141 0.229569527234482
142 0.229023521005045
143 0.228278341160457
144 0.2274909042367
145 0.227098021277093
146 0.226691522151918
147 0.226002251662269
148 0.226130072484178
149 0.227639667103224
150 0.227003931167295
151 0.227011928392586
152 0.226353226654079
153 0.225651809911107
154 0.224982215519177
155 0.225421965329807
156 0.226197020127695
157 0.225587437193919
158 0.224887189697497
159 0.224609712839495
160 0.232567631660877
161 0.232056687172895
162 0.23135701080315
163 0.23065841583017
164 0.230453446426468
165 0.229809551903973
166 0.229877901106919
167 0.229751715004421
168 0.229154094307905
169 0.228689465971559
170 0.229994771307647
171 0.23003744213933
172 0.229404259680804
173 0.228801733260573
174 0.228151711791114
175 0.228412990866297
176 0.228706717771197
177 0.228841071741695
178 0.229412273658371
179 0.228821727465187
180 0.22858018456467
181 0.227953403699319
182 0.228435233612799
183 0.228302136777916
184 0.22789783049733
185 0.227337656137426
186 0.226729005607494
187 0.226608967291566
188 0.226434627046778
189 0.227936324284063
190 0.22737786838667
191 0.22772813612408
192 0.22726999705628
193 0.227355678604077
194 0.226989888145998
195 0.226630590537491
196 0.226324592462519
197 0.225864297708627
198 0.225864297708627
199 0.225864297708627
};
\addlegendentry{black-box}
\addplot [semithick, darkorange25512714]
table {%
0 0
1 0.0489698133228855
2 0.0808273585885234
3 0.0700194488734825
4 0.150207855552065
5 0.144433835469471
6 0.225882499880619
7 0.215520288673953
8 0.208515047365098
9 0.200074810463183
10 0.230350641702421
11 0.222272576973733
12 0.214652800269648
13 0.217002645143545
14 0.211033867407235
15 0.205767933251995
16 0.201629043723523
17 0.196640245094096
18 0.191644138782006
19 0.187189096243669
20 0.188607555356115
21 0.193158976614295
22 0.189180432810254
23 0.185208415763116
24 0.181488526233988
25 0.180726698233356
26 0.185345309041014
27 0.194611634653141
28 0.193799611189774
29 0.19210073622923
30 0.198618930440591
31 0.196249617946758
32 0.193281644441073
33 0.195952939681306
34 0.193173795855153
35 0.193418225812629
36 0.192941526821293
37 0.213840193976271
38 0.211213654632038
39 0.208670064846094
40 0.207194123407834
41 0.206656302873375
42 0.20523052056137
43 0.203814114312931
44 0.204172086544691
45 0.203479691947627
46 0.201333760421184
47 0.199798351150014
48 0.197763996624052
49 0.219237681272789
50 0.217108197928336
51 0.216254300109322
52 0.215420253632676
53 0.213418656285873
54 0.21476950795451
55 0.213251459469028
56 0.214654807549877
57 0.212824989826303
58 0.214137116497076
59 0.212563517464029
60 0.211352094695762
61 0.215990687984366
62 0.214291442439776
63 0.213386391999141
64 0.22140952055853
65 0.220953428018616
66 0.219323849637877
67 0.217837875595797
68 0.217873590512554
69 0.219481600441045
70 0.218219437353331
71 0.216845316987508
72 0.215387278728306
73 0.215124466718316
74 0.214308068777974
75 0.213400539170926
76 0.212145370335166
77 0.217914872075614
78 0.216915689312951
79 0.216078161663976
80 0.214744224426098
81 0.2134385471703
82 0.212289638182711
83 0.211666254630243
84 0.211846873788455
85 0.210791570410045
86 0.211554748894203
87 0.210713259078811
88 0.209930472853118
89 0.209261124954433
90 0.208335544653264
91 0.208301897998221
92 0.208883901954101
93 0.208413316463197
94 0.207313548339727
95 0.208973812109127
96 0.20798856241785
97 0.20692467732386
98 0.207811201820056
99 0.207349015109965
100 0.207343697358728
101 0.206350022862676
102 0.205346416707042
103 0.204528386022341
104 0.203696259766376
105 0.202850703655285
106 0.202059993415613
107 0.202361055637206
108 0.209130210603815
109 0.216185983414408
110 0.216563337511767
111 0.215662818014897
112 0.216535202123056
113 0.215628921937563
114 0.215951350591704
115 0.215194656955364
116 0.215716254584673
117 0.217195360565846
118 0.217147774730994
119 0.218627789691608
120 0.218480516190754
121 0.218580948835081
122 0.217691102688061
123 0.216821367381364
124 0.216193650450132
125 0.2162733167188
126 0.215469133017169
127 0.215138266461948
128 0.215014577652469
129 0.215147427699305
130 0.21439897102656
131 0.215633456463324
132 0.214865997293974
133 0.245114138346918
134 0.244627776327837
135 0.24425740149672
136 0.243505106148772
137 0.242917150416216
138 0.242597684591089
139 0.241990010872559
140 0.241188127624934
141 0.24036482130824
142 0.240479508895233
143 0.240433965511705
144 0.23969697456839
145 0.239486728975755
146 0.24038644658855
147 0.24095151204543
148 0.240231485068076
149 0.240126688200477
150 0.240082132397253
151 0.239291192053095
152 0.238604418832219
153 0.23846500723982
154 0.237954531753492
155 0.23720862505227
156 0.239517539729977
157 0.238772061722061
158 0.238473444908047
159 0.239323306412811
160 0.240270608388823
161 0.239733176581927
162 0.240597765428016
163 0.240583640628986
164 0.240110859790156
165 0.239458252653447
166 0.23920078935472
167 0.238519839453217
168 0.238614601499619
169 0.23905537067738
170 0.238400923774456
171 0.238508340592655
172 0.237956444818345
173 0.237709238844831
174 0.237270085612087
175 0.236605239263249
176 0.236173904370825
177 0.235509571337441
178 0.234852320624898
179 0.234530870116973
180 0.233882211868592
181 0.2341957748218
182 0.233557117411979
183 0.232940936628163
184 0.233259870087121
185 0.232732987895422
186 0.232385788425639
187 0.231824582620521
188 0.231476651673705
189 0.231239487540872
190 0.230755554770434
191 0.230167664504077
192 0.230180610884661
193 0.232066773408142
194 0.231510413034661
195 0.23114996247374
196 0.230644721515839
197 0.230098412745047
198 0.232138802750635
199 0.231657929908777
};
\addlegendentry{grey-box}
\end{axis}

\end{tikzpicture}
    \label{fig:StdHistory}}
    \caption{The figure shows the long-term convergence of the online optimization algorithm when employing either of the two approaches to deriving a landing point prediction model. Figure (a) shows the mean distance error $\epsilon_i$ with a graph $\mathcal{O}(i^{-\frac{1}{2}})$ (scaled to match $\epsilon_1$) for reference and figure (b) shows the standard deviation $\sigma_i$ of landing points. The results in (a) and (b) originate from the same experiment, where step lengths defined by $\alpha^1 = 0.05$ were used.}
    \label{fig:Robustness}
\end{figure}

In Figure~\ref{fig:Robustness}, we quantitatively show the evolution of the distance error $\epsilon_i$ and the standard deviation $\sigma_i$ as defined in~\eqref{eq:PerformanceMetrics}, respectively. We note that the grey-box and the black-box gradient approximations result in similar convergence rates. This implies that our online learning algorithm is robust to modeling errors. Furthermore, Figure~\ref{fig:StdHistory} highlights that the standard deviation $\sigma_i$ of our algorithm has converged to $\underline{\sigma}$ (the robot's inherent variance).
Each run with $200$ iterations takes roughly $30$ minutes in our setup, whereby no ball was missed.

\subsection{Convergence}
\label{sec:ResultsRobustness}
In the following experiments we showcase the convergence properties of our algorithm with the black-box model for varying settings in terms of different target choices (Figure~\ref{fig:Targets}) and different initial guesses for the policy (Figure~\ref{fig:InitialGuesses}).
Here we increase the step lengths ($\alpha^1 = 0.15$) to emphasise convergence speed. In conclusion, the results of our experiments demonstrate the rapid convergence of our algorithm (about $2$-$5$ iterations) under various conditions, showcasing its efficiency and effectiveness in learning interception policies.

\begin{figure}
    \centering
    \subfloat[Evolution of distance error]{
    \hspace{-5pt}
\begin{tikzpicture}

\definecolor{crimson2143940}{RGB}{214,39,40}
\definecolor{darkgray176}{RGB}{176,176,176}
\definecolor{darkorange25512714}{RGB}{255,127,14}
\definecolor{forestgreen4416044}{RGB}{44,160,44}
\definecolor{lightgray204}{RGB}{204,204,204}
\definecolor{mediumpurple148103189}{RGB}{148,103,189}
\definecolor{sienna1408675}{RGB}{140,86,75}
\definecolor{steelblue31119180}{RGB}{31,119,180}

\begin{axis}[
height=0.5\linewidth,
width=\linewidth, 
legend columns = 3,
legend cell align={left},
legend style={
  fill opacity=0.8,
  draw opacity=1,
  text opacity=1,
  at={(0.987,0.97)},
  draw=lightgray204
},
tick align=outside,
tick pos=left,
x grid style={darkgray176},
xlabel={iteration \(\displaystyle i\) [-]},
xmin=-0.95, xmax=19.95,
xtick style={color=black},
y grid style={darkgray176},
ylabel={distance error [m]},
ymin=-0.0602893616972287, ymax=1.59109329666393,
ytick style={color=black},
ylabel shift = 3 pt
]
\addplot [steelblue31119180, mark = *, mark size = 1pt,  mark options = solid]
table {%
0 1.17398357158786
1 0.372592089513412
2 0.518243814542441
3 0.209758257714097
4 0.458414253699673
5 0.290283044308836
6 0.482544473144308
7 0.179717684211681
8 0.233327855123336
9 0.30631290771178
10 0.156650425043632
11 0.316248990674689
12 0.339735245810032
13 0.277330828942242
14 0.307934111191405
15 0.278524693949447
16 0.258057668849206
17 0.0516716751467983
18 0.45488721522277
19 0.145524959486459
};
\addlegendentry{target 1}
\addplot [darkorange25512714, mark = *, mark size = 1pt,  mark options = solid]
table {%
0 1.10003973450235
1 0.410106042901479
2 0.371675506344746
3 0.0989227617263049
4 0.339237842068554
5 0.0426875974290113
6 0.270753601280746
7 0.0906048246684691
8 0.0901608306698492
9 0.107919323238194
10 0.125659059215087
11 0.0975502201254612
12 0.240326454410574
13 0.224432078963486
14 0.113015579024648
15 0.617938900724918
16 0.144083910291739
17 0.291720647024114
18 0.376441572187797
19 0.159813695007748
};
\addlegendentry{target 2}
\addplot [forestgreen4416044, mark = *, mark size = 1pt,  mark options = solid]
table {%
0 1.5160304485566
1 0.353662184641884
2 0.140348013854936
3 0.392283573830044
4 0.319577158749444
5 0.141861769723268
6 0.232523588412406
7 0.0767398974790204
8 0.23582678963869
9 0.0565785215753525
10 0.284424399065831
11 0.268195572677732
12 0.0634763166197526
13 0.225965929329295
14 0.323349100639612
15 0.0721684582081238
16 0.140077381929713
17 0.16573275134913
18 0.197949974859932
19 0.420810732222281
};
\addlegendentry{target 3}
\addplot [crimson2143940,
mark = *,
mark size = 1pt,  mark options = solid]
table {%
0 0.457148494154416
1 0.18117220883005
2 0.108886107065156
3 0.144577558324448
4 0.242887901571714
5 0.101581895330019
6 0.103397459444964
7 0.246720110670356
8 0.187493913933563
9 0.212048915059497
10 0.068690804888077
11 0.107685804166545
12 0.191785998016584
13 0.029949042455466
14 0.0899880608413725
15 0.0323538627246877
16 0.128665957085342
17 0.16946874297136
18 0.110746548203549
19 0.118832689371285
};
\addlegendentry{target 4}
\addplot [mediumpurple148103189,
mark = *,
mark size = 1pt,  mark options = solid]
table {%
0 0.302414131465507
1 0.0147734864100965
2 0.302853708385479
3 0.0964889441856589
4 0.103537046811052
5 0.0954644542519977
6 0.315972411907161
7 0.14976900018697
8 0.177468771472753
9 0.134524582175633
10 0.0991861738553547
11 0.13894427936431
12 0.145633827996566
13 0.25878289334065
14 0.218634922733225
15 0.0206609879259138
16 0.104184756337826
17 0.0268117366690643
18 0.252818362373383
19 0.089112271778246
};
\addlegendentry{target 5}
\addplot [sienna1408675,
mark = *,
mark size = 1pt,  mark options = solid]
table {%
0 0.395995435996351
1 0.263437678299775
2 0.303467902095577
3 0.167976011962485
4 0.136442807594045
5 0.207992816487587
6 0.195297513972914
7 0.289457072538498
8 0.105813096910628
9 0.346217179006023
10 0.0873285098061261
11 0.192040494180454
12 0.0570298732337305
13 0.102827176975262
14 0.0525383243146211
15 0.135053943372499
16 0.12682903156367
17 0.100262491692558
18 0.0957338818466193
19 0.043071575473892
};
\addlegendentry{target 6}
\end{axis}

\end{tikzpicture} \centering
    \label{fig:Targets_ErrorHistory}}\\
    \subfloat[Evolution of policy]{\centering
\begin{tikzpicture}

\definecolor{crimson2143940}{RGB}{214,39,40}
\definecolor{darkgray176}{RGB}{176,176,176}
\definecolor{darkorange25512714}{RGB}{255,127,14}
\definecolor{forestgreen4416044}{RGB}{44,160,44}
\definecolor{lightgray204}{RGB}{204,204,204}
\definecolor{mediumpurple148103189}{RGB}{148,103,189}
\definecolor{sienna1408675}{RGB}{140,86,75}
\definecolor{steelblue31119180}{RGB}{31,119,180}

\begin{groupplot}[group style={group size=1 by 2, vertical sep=1cm}, 
height=0.45\linewidth,
width=\linewidth]
\nextgroupplot[
tick align=outside,
tick pos=left,
x grid style={darkgray176},
xmin=-0.95, xmax=19.95,
xtick style={color=black},
y grid style={darkgray176},
ylabel={\(\displaystyle \theta_1(t_{\text{ic}})\) [deg]},
ymin=2.23765991601051, ymax=29.0044233163369,
ytick style={color=black},
ylabel shift = 5 pt,
xtick={-9, -8}
]
\addplot [mark = *, mark size = 1pt,  mark options = solid, semithick, steelblue31119180]
table {%
0 11.4591559026165
1 19.857744029207
2 22.6884412147654
3 21.4586335918451
4 22.6503846728824
5 23.5785726212924
6 23.9767920757696
7 23.9336053056578
8 24.8405872448365
9 25.0603548132608
10 24.9268997300425
11 24.604931157609
12 25.6333771966
13 26.7402466331712
14 27.2387965877132
15 27.4222977432179
16 26.9638650015624
17 27.6993180135996
18 27.7877522526857
19 27.3483237153713
};
\addplot [mark = *, mark size = 1pt,  mark options = solid, semithick, darkorange25512714]
table {%
0 11.4591559026165
1 19.8820398317776
2 14.1912073246087
3 13.5251264906463
4 14.704819294024
5 14.7141307637975
6 14.3331943090159
7 12.3548738746302
8 12.0834600542312
9 12.8348464968925
10 12.4930519298029
11 12.9496287245646
12 13.666511818262
13 13.0282247802354
14 12.3509090027642
15 13.049569540945
16 12.1843337505172
17 13.1049421155182
18 13.2484749048376
19 13.353682746541
};
\addplot [mark = *, mark size = 1pt,  mark options = solid, semithick, forestgreen4416044]
table {%
0 11.4591559026165
1 7.04759331353226
2 3.45433097966171
3 5.5506727315663
4 6.8200047334319
5 6.3292665741979
6 6.7302528946575
7 4.69534395627235
8 4.19646127217829
9 5.8354914320769
10 5.29929840079488
11 6.83161841846763
12 6.15770943679259
13 5.93109397428401
14 5.86670783018308
15 5.80884742533557
16 5.80049141805779
17 4.95468179403639
18 3.63595992541173
19 4.34778630181789
};
\addplot [mark = *, mark size = 1pt,  mark options = solid, semithick, crimson2143940]
table {%
0 11.4591559026165
1 18.559226117274
2 19.8835800424585
3 20.5342984669142
4 21.0903572663984
5 23.03004402522
6 22.4616543548468
7 23.1488064354527
8 22.1999118223962
9 23.2508406387277
10 22.1722261815996
11 21.8745088890823
12 21.3442271389493
13 22.1716472431892
14 22.0326786423779
15 22.3772715993493
16 22.393783370972
17 22.7925789697646
18 22.4145842818542
19 22.0864896372831
};
\addplot [mark = *, mark size = 1pt,  mark options = solid, semithick, mediumpurple148103189]
table {%
0 11.4591559026165
1 11.4591559026165
2 11.5754875074179
3 9.47493744129652
4 8.65777998579151
5 8.6936603644962
6 8.27613810542778
7 9.19707369269761
8 8.3554641589852
9 7.90556555994812
10 7.50006195568408
11 7.25035932596085
12 7.75305057497218
13 7.17125865557949
14 7.85886449734654
15 8.58161880968489
16 8.5778143178668
17 8.69335032659996
18 8.72359255821129
19 7.86236371127078
};
\addplot [mark = *, mark size = 1pt,  mark options = solid, semithick, sienna1408675]
table {%
0 11.4591559026165
1 6.0838363292645
2 8.70837175511737
3 7.95247137625908
4 6.61299241282948
5 5.76698923524732
6 7.14801420115533
7 5.9978715902654
8 6.54266105963313
9 6.16651300431274
10 6.82895416762074
11 7.29041213762395
12 7.11017223371581
13 7.11230043480096
14 7.16004414056738
15 7.16678484813553
16 7.6379138570604
17 8.09273328467175
18 7.73626008736483
19 7.828990252621
};

\nextgroupplot[
tick align=outside,
tick pos=left,
x grid style={darkgray176},
xlabel={iteration \(\displaystyle i\) [-]},
xmin=-0.95, xmax=19.95,
xtick style={color=black},
y grid style={darkgray176},
ylabel={\(\displaystyle \theta_4(t_{\text{ic}})\) [deg]},
ymin=-21.167605075025, ymax=8.82886206747196,
ytick style={color=black},
ylabel shift = -3 pt,
yshift=0.5cm,
]
\addplot [mark = *, mark size = 1pt,  mark options = solid, semithick, steelblue31119180]
table {%
0 -14.3239448782706
1 -5.32846600129068
2 -3.75757499329245
3 -2.05179533147441
4 -1.4737006275917
5 0.0302216848909558
6 1.0320403308344
7 2.72224562934731
8 2.97280074900342
9 2.14646338269274
10 3.14492030043026
11 3.58199175162425
12 3.00153140514958
13 3.65245101970649
14 4.46837257125345
15 5.45510396644303
16 4.65534236047785
17 5.05517099560067
18 4.94700419361921
19 3.66251519427763
};
\addplot [mark = *, mark size = 1pt,  mark options = solid, semithick, darkorange25512714]
table {%
0 -14.3239448782706
1 0.592863745623403
2 -1.75486520616551
3 -0.0723207996016473
4 -0.337613067353875
5 1.08860461735261
6 0.976024784120402
7 1.90263864923303
8 1.57086715843463
9 1.69244737061616
10 2.07383182325358
11 2.45267242647306
12 2.59812574697486
13 3.50411788583262
14 2.70558700290566
15 2.42615552882585
16 0.529835763156476
17 0.380871637607119
18 -0.269902550468977
19 0.482700161754979
};
\addplot [mark = *, mark size = 1pt,  mark options = solid, semithick, forestgreen4416044]
table {%
0 -14.3239448782706
1 0.765621712181983
2 2.18882053522197
3 1.82460471271465
4 2.65511421163061
5 3.67151247765907
6 7.46538628826756
7 3.89202509417604
8 4.17337662891156
9 4.07879220294237
10 4.19555413372201
11 4.33343494365237
12 5.16269201174905
13 5.04717375945916
14 5.66949426247579
15 6.64161467302747
16 6.89276285722131
17 6.71012822667621
18 6.86607468798594
19 7.16077434358136
};
\addplot [mark = *, mark size = 1pt,  mark options = solid, semithick, crimson2143940]
table {%
0 -14.3239448782706
1 -12.3849560288711
2 -13.0094377073366
3 -12.2793397423012
4 -12.7601453206179
5 -12.1770838714251
6 -12.6435131886073
7 -12.5035913521386
8 -13.5295297626654
9 -13.3504504122515
10 -14.0107257958642
11 -13.9992028431575
12 -14.1341495828984
13 -13.6195683888352
14 -13.6565974459086
15 -13.4129913121041
16 -13.4884407027736
17 -13.133862776133
18 -13.6057058528298
19 -13.5458359701174
};
\addplot [mark = *, mark size = 1pt,  mark options = solid, semithick, mediumpurple148103189]
table {%
0 -14.3239448782706
1 -14.3239448782706
2 -14.3748673518312
3 -16.664745803322
4 -16.9193111175438
5 -17.3881807155899
6 -17.2458213878583
7 -17.9751785029548
8 -18.2185748801957
9 -18.911106266159
10 -19.382858208576
11 -19.2374801841755
12 -19.2600616170564
13 -19.5627712207565
14 -19.7961946752749
15 -19.7869096129731
16 -19.738886490228
17 -19.4657741768771
18 -19.512582336733
19 -19.8041292958206
};
\addplot [mark = *, mark size = 1pt,  mark options = solid, semithick, sienna1408675]
table {%
0 -14.3239448782706
1 -14.2157733182714
2 -12.9361624143614
3 -11.567066628156
4 -11.7249986849687
5 -11.6757023959817
6 -11.2777706688009
7 -11.3629259478863
8 -10.3197065225049
9 -10.1752266091449
10 -9.06277713329763
11 -8.88340417488655
12 -8.44598941588707
13 -8.58380947549425
14 -8.33092574803073
15 -8.44743706847793
16 -8.13223555760395
17 -8.18300051487159
18 -8.39688542393753
19 -8.5615432744357
};
\end{groupplot}

\end{tikzpicture}
    \label{fig:Targets_PolicyEvolution}}\\
    \subfloat[Last five landing points]{\centering
\begin{tikzpicture}

\definecolor{crimson2143940}{RGB}{214,39,40}
\definecolor{darkgray176}{RGB}{176,176,176}
\definecolor{darkorange25512714}{RGB}{255,127,14}
\definecolor{forestgreen4416044}{RGB}{44,160,44}
\definecolor{lightgray204}{RGB}{204,204,204}
\definecolor{mediumpurple148103189}{RGB}{148,103,189}
\definecolor{sienna1408675}{RGB}{140,86,75}
\definecolor{steelblue31119180}{RGB}{31,119,180}

\begin{axis}
[
width=\linewidth,
legend cell align={left},
legend columns = 3,
legend style={
  fill opacity=0.8,
  draw opacity=1,
  text opacity=1,
  at={(0.5,1.05)},
  anchor=south,
  draw=lightgray204
},
tick align=outside,
tick pos=left,
x grid style={darkgray176},
xlabel={landing point $x$ [m]},
xmin=-1.224, xmax=0.876,
xtick style={color=black},
y grid style={darkgray176},
ylabel={landing point $y$ [m]},
ymin=1.2, ymax=3.7,
ytick style={color=black}
]
\draw[draw=black,draw opacity=0.5,dashed, fill=none] (axis cs:-0.174-1.525/2, 1.63-2.74/2) rectangle (axis cs:-0.174+1.525/2, 1.63+2.74/2);
\addlegendimage{ybar, draw=black, draw opacity=0.5, dashed, area legend}
\addlegendentry{table}
\draw[dashed] (-0.174 - 1.7/2, 1.63) -- (-0.174 + 1.7/2, 1.63);

\draw[rotate around={30:(-0.75, 2.86)}] (-0.75, 2.86) ellipse (0.1089 and 0.22684);
\draw (-0.17, 2.86) ellipse (0.1089 and 0.22684);
\draw[rotate around={-10:(0.41, 2.86)}] (0.41, 2.86) ellipse (0.1089 and 0.22684);
\draw[rotate around={20:(-0.46, 1.95)}] (-0.46, 1.95) ellipse (0.10427 and 0.07404);
\draw (-0.17, 1.75) ellipse (0.10427 and 0.07404);
\draw (0.12, 1.95) ellipse (0.10427 and 0.07404);

\addplot[draw=black, mark=+, mark size=5pt, very thick] (10, 10);
\addlegendentry{targets}
\addplot[draw=black, mark=*, only marks,opacity=0.6] (10, 10);
\addlegendentry{landing points}

\addplot [
  draw=steelblue31119180,
  fill=steelblue31119180,
  forget plot,
  mark=*,
  only marks,
  opacity=0.6
]
table{%
x  y
-0.833001420844067 3.12586983521547
-0.54403835300583 2.70452151139358
-0.709394394401594 2.89195538775329
-0.790171188124042 3.3131099802673
-0.627775642218681 2.78101443044151
};
\addplot [draw=none, draw=steelblue31119180, fill=steelblue31119180, forget plot, mark=+, mark size=5pt, very thick]
table{%
x  y
-0.75 2.86
};
\addplot [
  draw=darkorange25512714,
  fill=darkorange25512714,
  forget plot,
  mark=*,
  only marks,
  opacity=0.8
]
table{%
x  y
-0.0562049497316959 3.46737070357695
-0.0262370281705325 2.86961151058458
-0.0713752072587711 3.13454341397476
-0.243768201002695 2.49085706048734
-0.316375319134874 2.79585420465694
};
\addplot [draw=darkorange25512714, draw=none, fill=darkorange25512714, forget plot, mark=+, mark size=5pt, very thick]
table{%
x  y
-0.17 2.86
};
\addplot [
  draw=forestgreen4416044,
  fill=forestgreen4416044,
  forget plot,
  mark=*,
  only marks,
  opacity=0.8
]
table{%
x  y
0.354792690206426 2.81351947391117
0.431206277860735 2.99846287122393
0.357514820777803 3.01720257896029
0.32626304839965 2.68063357202718
0.348871599590568 2.44365279991714
};
\addplot [draw=forestgreen4416044, draw=none, fill=forestgreen4416044, forget plot, mark=+, mark size=5pt, very thick]
table{%
x  y
0.41 2.86
};
\addplot [draw=crimson2143940, fill=crimson2143940, forget plot, mark=*, only marks, opacity=0.8]
table{%
x  y
-0.453419638346923 1.98167761471012
-0.374978804322996 1.85342710111869
-0.53844648363013 2.10021918669185
-0.547887480653567 1.88261610961536
-0.552844128878057 2.02416991166293
};
\addplot [draw=crimson2143940, draw=none, fill=crimson2143940, forget plot, mark=+, mark size=5pt, very thick]
table{%
x  y
-0.46 1.95
};
\addplot [
  draw=mediumpurple148103189,
  fill=mediumpurple148103189,
  forget plot,
  mark=*,
  only marks,
  opacity=0.8
]
table{%
x  y
-0.179921738391476 1.73187720966943
-0.183990071679378 1.64675881951674
-0.151068538956626 1.76898602133082
-0.40279601740159 1.84860597667037
-0.253979971242604 1.77980539232373
};
\addplot [draw=mediumpurple148103189, draw=none, fill=mediumpurple148103189, forget plot, mark=+, mark size=5pt, very thick]
table{%
x  y
-0.17 1.75
};
\addplot [draw=sienna1408675, fill=sienna1408675, forget plot, mark=*, only marks, opacity=0.8]
table{%
x  y
0.1754545052561 1.82685632185425
0.24299739242174 1.98093937143561
0.0696841401304956 2.03672301589539
0.177445460615608 2.02658325657794
0.0796824717373013 1.96515445573386
};
\addplot [draw=none, draw=sienna1408675, fill=sienna1408675, forget plot, mark=+, mark size=5pt, very thick]
table{%
x  y
0.12 1.95
};
\end{axis}

\end{tikzpicture}
    \label{fig:Targets_LandingPoints}}\\
    \caption{The figure shows the results when applying the method for six different targets, each with the same initial guess. In (a) we see the evolution of the distance error, in (b) we see the evolution of the policy parameters $\theta_1(t_{\text{ic}})$ and $\theta_4(t_{\text{ic}})$, and in (c) we see the last five individual landing points corresponding to each target. The gradient approximation is done using the black-box model. Step lengths defined by $\alpha^1 = 0.15$ were used.}
    \label{fig:Targets}
\end{figure}

\begin{figure}[h]
    \centering
    \vspace*{11pt}
    {\centering
\begin{tikzpicture}

\definecolor{darkgray176}{RGB}{176,176,176}
\definecolor{gray}{RGB}{128,128,128}
\definecolor{lightgray204}{RGB}{204,204,204}

\begin{axis}[
height=0.7\linewidth,
width=\linewidth,
legend cell align={left},
legend columns = 2,
legend style={
  fill opacity=0.8,
  draw opacity=1,
  text opacity=1,
  at={(0.5,1.05)},
  anchor=south,
  draw=lightgray204
},
log basis y={10},
tick align=outside,
tick pos=left,
x grid style={darkgray176},
tick align=outside,
tick pos=left,
x grid style={darkgray176},
xlabel={\(\displaystyle y\) [m]},
ymin=-0.6215-0.55, ymax=0.6215+0.55,
xtick style={color=black},
xtick={1,2,3},
y grid style={darkgray176},
ylabel={\(\displaystyle x\) [m]},
xmin=0.37-0.2, xmax=3.11+0.7,
ytick style={color=black},
y label style={at={(axis description cs:-0.1,.5)},anchor=south}
]

\addplot [draw=black, draw=none, fill=none, mark=+, only marks]
table{%
x  y
-10 0
};
\addlegendentry{initial guesses}

\addplot [draw=black, fill=black, mark=*, mark size=1pt]
table {%
-2.28958353011993 0.858212607100258 
-2.77740886241278 0.925857660245247 
};
\addlegendentry{landing point history}

\addplot [thick, draw=black, fill=black,  mark=x, only marks, mark size = 4pt]
table{%
x  y
 2.7855699 0.16021162 
};
\addlegendentry{target}

\draw[draw=black,draw opacity=0.5,dashed, fill=none] (axis cs:0.37,-0.6215) rectangle (axis cs:3.11,0.9035);
\addlegendimage{ybar, draw=black, draw opacity=0.5, dashed, area legend}

\addplot [draw=black, draw=none, fill=none, mark=+, only marks]
table{%
x  y
1.81775371394647 0.762360828231538 
3.15175041450899 0.705103160497393 
3.69841192523518 -0.334733830573781 
2.79797749605647 -1.04408066641058 
2.51774586176497 -0.836461580621605 
1.91180103358081 0.133500722756789 
};

\addplot [semithick, gray, forget plot]
table {%
1.81775371394647 0.762360828231538 
2.7587976587379 0.7296956480599135
2.9098581544751005 0.31719263191373415
2.811288214813606 0.09401007328250251
2.8495990007238783 0.11463692805951979
};
\addplot [semithick, gray, forget plot]
table {%
3.15175041450899 0.705103160497393 
2.9751679847559496 -0.09749058575103466
2.834607164570688 0.07988239021925085
2.7664379829168375 0.030591196952036336
2.7609589346008976 0.005866919805501833
};
\addplot [semithick, gray, forget plot]
table {%
3.69841192523518 -0.334733830573781 
3.2229608142389106 -0.17350108705013484
2.8942439287559636 -0.20727974873080557
2.8885832535058555 0.0015729227929873108
2.656804602422089 0.03419613627690195
};
\addplot [semithick, gray, forget plot]
table {%
2.79797749605647 -1.04408066641058 
2.753834516876263 -0.4755105288924506
2.682896213057152 -0.4111119940086968
2.678199592569998 -0.22383507747937478
2.651464484052073 -0.020873320777696203
};
\addplot [semithick, gray, forget plot]
table {%
2.51774586176497 -0.836461580621605 
2.4268029321361015 -0.37785256207143475
2.4930585197560844 -0.24702285192504098
2.576042121790235 -0.11097510867230757
2.8440938701130087 0.17029861200855467
};
\addplot [semithick, gray, forget plot]
table {%
1.91180103358081 0.133500722756789 
2.55587698097398 0.4035728486516808
2.8567728573737328 0.38260845895312495
2.623117883525976 0.28877716672106285
2.8172769948422647 0.2917340577401453
};

\addplot [draw=black, fill=black, forget plot, mark=*, only marks, mark size=1pt]
table{%
x  y
2.7587976587379 0.7296956480599135
2.9098581544751005 0.31719263191373415
2.811288214813606 0.09401007328250251
2.8495990007238783 0.11463692805951979

2.9751679847559496 -0.09749058575103466
2.834607164570688 0.07988239021925085
2.7664379829168375 0.030591196952036336
2.7609589346008976 0.005866919805501833

3.2229608142389106 -0.17350108705013484
2.8942439287559636 -0.20727974873080557
2.8885832535058555 0.0015729227929873108
2.656804602422089 0.03419613627690195

2.753834516876263 -0.4755105288924506
2.682896213057152 -0.4111119940086968
2.678199592569998 -0.22383507747937478
2.651464484052073 -0.020873320777696203

2.4268029321361015 -0.37785256207143475
2.4930585197560844 -0.24702285192504098
2.576042121790235 -0.11097510867230757
2.8440938701130087 0.17029861200855467

2.55587698097398 0.4035728486516808
2.8567728573737328 0.38260845895312495
2.623117883525976 0.28877716672106285
2.8172769948422647 0.2917340577401453
};

\end{axis}

\end{tikzpicture}
    \label{fig:InitialGuesses}}
    \caption{The figure shows the first five landing points for different initial interception policies $\phi^{1}$. Each landing point history corresponds to an average over five individual runs for any given initial policy. The gradient approximation is done using the grey-box model. Step lengths defined by $\alpha^1 = 0.1$ were used.}
    \label{fig:InitialGuesses}
\end{figure}
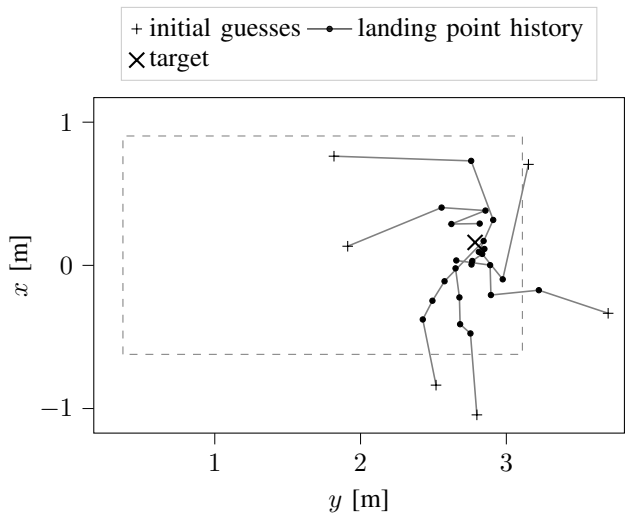

\section{Conclusion}\label{sec:Conclusion}

In summary, the article proposes a robust and data-efficient online optimization algorithm and successfully applies it to return incoming balls to a predefined target with a table tennis robot. We compare two landing point prediction models based on either a grey-box or a black-box approach, which are used to approximate the gradient in our optimization framework. We compare the long-term ($200$ iterations) policy learning of the framework and find that in both cases we converge at similar rates. This highlights the robustness of our algorithm. Starting from a wide range of initializations we can hit targets with close to zero mean and a standard deviation of about \SI{25}{\cm}. The standard deviation of \SI{25}{\cm} corresponds to the inherent non-repeatability due to the variance in incoming ball trajectories and the process noise of our robot, which is driven by soft pneumatic actuators. By applying our online optimization algorithm, the landing point converges, starting from any initial policy, quickly to a predefined target on the table taking $2$-$5$ iterations to reach an accuracy of about \SI{25}{\cm}.


\section*{ACKNOWLEDGMENT}
We thank the German Research Foundation, the Branco Weiss Fellowship, administered by ETH Zurich, and the Center for Learning Systems for the support.

\bibliographystyle{IEEEtran} 
\bibliography{IEEEabrv,bibliography}

\end{document}